# Answer Extraction for Why Arabic Questions Answering Systems: EWAQ

Fatima T. AL-Khawaldeh
Department of Computer Science
Al-Albayt University
Al-Mafraq, Jordan

Abstract —with the increasing amount of web information, questions answering systems becomes very important to allow users to access to direct answers for their requests. This paper presents an Arabic Questions Answering Systems based on entailment metrics. The type of questions which this paper focuses on is why questions. There are many reasons lead us to develop this system: generally, the lack of Arabic Questions Answering Systems and scarcity Arabic Questions Answering Systems which focus on why questions. The goal of the proposed system in this research is to extract answers from re-ranked retrieved passages which are retrieved by search engines. This system extracts the answer only to why questions. This system is called by EWAQ: Entailment based Why Arabic Questions Answering. Each answer is scored with entailment metrics and ranked according to their scores in order to determine the most possible correct answer. EWAQ is compared with search engines: yahoo, google and ask.com, the well-established web-based Questions Answering systems, using manual test set. In EWAQ experiments, it is showed that the accuracy is increased by implementing the textual entailment in re-raking the retrieved relevant passages by search engines and deciding the correct answer. The obtained results show that using entailment based similarity can help significantly to tackle the why Answer Extraction module in Arabic language.

## I. INTRODUCTION

Recently, the huge amount of information on web increased the need to retrieve the exact answer rather than several documents. Question Answering (QA) is one of the most important fields in Information Retrieval (IR) which helps users to get precise information unlike the search engines which retrieve web pages. QA systems consists of three main components question processing, IR, and Answer Extraction (AE). Question processing receives the user request, classifies the question to its type and determine the expected answers including name entities. There are three main types of questions as classified in [1]: factoid, yes/no, why, definition, each type has different approach. IR module retrieves the related web documents. AE module depends on name entity recognizer to retrieve all the possible answers (candidate's answers) which have the information type. For example, if we have this Arabic question:" eaina taqqa Amman?" "اين تقع عمان؟", the expected type of the answer is place, so all the places will be received. The name entity recognizer distinguishes the places as entity type.

There are two main types of QA systems: open-domain QA systems and closed domain QA systems. Open-domain QA systems concern of everything unlike closed domain QA systems which concentrate on specific domain like (computer, energy, music etc.).

Many evaluation metrics were discussed in previous researches, judge the answer based on some criteria like the relevancy, correctness, Conciseness (not contains irrelevant information), and Completeness (not part of answer) [2]. Retrieving short answers is the main challenge of most QA systems [3].There are several evaluation metrics comes from TREC, CLEF, NTCIR, etc.: Precision, recall and F-measure where:

Precision =number of correct answers /number of questions answered           (1)

Recall =number of correct answers /number of questions to be answered           (2)

F measure =2* (Precision * recall) /Precision  +recall (3) [3].

More Arabic speakers and increased amount of Arabic content on the internet increased the need to develop

Arabic QA system able to retrieve precise answer rather than full documents. Due to challenges of Arabic language, the performance of AQAS is less than English and other Latin language.

*CHALLENGES OF ARABIC LANGUAGE*

Arabic languages has many challenges:

- Complex morphology.
- A lack of natural language processing tools and resources (corpora, dictionaries, lexicon etc.).
- Highly derivational and inflectional.
- Ambiguity (different meaning of the same word).
- Not including capital letters which distinguishes the names from verbs.

Since the why QA systems need more processing, more complex artificial intelligent and complex natural language processing techniques, most previous AQAS excluded it from their researches. In this paper, we try to focus on the why questions in Arabic language using the entailment metrics. Often, the answer type of these questions is reason.

The rest of the paper is organized to sections: Section II presents related works of QA systems, Section III describes the entailment relation, Section IV presents entailment adaptation in Arabic QA system (EWAQ), and Section V presents the experiments conducted in EWAQ system and discusses the obtained results which presented in this section. The main conclusions are showed in Section VI.

## II. RELATED WORKS

There are a lots of important Latin QA systems, we presents some of them:

- QALC system is popular QA system for English language. This system developed in the text retrieval conference (TREC) depending on syntactic and semantic analysis [4].
- QRISTAL system developed for more language (Italian, English, French, polish and Portuguese), it used the name entity recognizer to retrieve the precise answers from the internet [5].
- Ask.com is a well-known web based QA system, developed for Arabic and English language. Keywords is the key of the system to retrieve the answers [6].

For Arabic language, this paper presents the most important systems:

- AQSA is knowledge-based QA system that extracts answers from structured data but not published evaluation [7].
- QARAB is an Arabic QA systems that uses IR and NLP techniques to extract answers. It supports the factoid but excluded how and why [8].

- ArabiQA is Arabic QA system that concerns with factoid Questions that uses Named Entity Recognition techniques and Java IR System (JIRS) for Arabic text [9].
- QASAL is Arabic QA system for factoid questions which uses the NooJ platform [10] but not fully automatic. It uses IR and NLP techniques to extract the precise answers which requested by users. Published work based on this systems does not include experimental results or performance metrics [11].
- ArQA is Arabic QA system that deals answers to factoid questions system uses IR and NLP techniques. ArQA consists four modules: question processing, passage retrieval, AE and answer validation modules [12].
- The authors of [13] presented a new approach and a new QA handles all types of questions including (How and why), QArabPro is the name of their system. The overall accuracy of the system is 84% and accuracy of answers to the why questions is the lowest one 62, 22%. The researchers tried to found more keywords to get better results in this type of questions (why).
- In 2013, the authors of [14] showed that using a semantic logical representation based query expansion (synonyms and antonyms) improves the precision of the yes/no questions system.

## III. THE ENTAILMET RELATION

Text entailment method analyzes a text and provides different semantic inferences to enhance the results of the output. One of the main inferences is the degree of semantic connectedness among the parts of the text. The authors of [15] defined textual entailment as "an attempt to promote an abstract generic task that captures major semantic inference needs across applications". Several researches in NLP fields such as: IR, QA, Information Extraction (IE), and Text summarization (TS) use text entailment relation to accomplish these tasks.

Textual Entailment is determining if a statement is entailed by another statement or not. In order to check the entailment relation between tow texts, deep semantic analysis is needed. In EWAQ system, It is supposed that the Answer (A) is the statement which entails and Question (Q) is the entailed by statement A → Q. Using Wordnet to identify semantic relations helps understanding the links between the different parts of the text, then select sentences based on the semantic content of the sentence and the relative importance of the content to the semantics of the text [16]. Arabic WordNet (AWN) open source software is used in EWAQ to perform the pre-processing stage in order to find the related words to each word in the question and retrieved passage [17].

## IV. ENTAILMENT ADAPTATION IN ARABIC WHY QA SYSTEM (EWAQ)

The main components of QA systems are: Question Analysis, Passage Retrieval and AE. After identifying the expected type of answer in Question Analysis module and forming a query, the formed query is passed to Passage Retrieval to retrieve relevance passages. The most relevant passages are ranked according to its relevancy in Passage Retrieval module. The correct answer is obtained from the candidates' passages in AE module. In this work, the objective is enhancing the AE module and improving re-ranking of returned passages in PR modules.

As mentioned previously, the type of question implemented in EWAQ is why questions. Search engines are used in EWAQ system in order to retrieve passages which will be re-ranked by entailment relation. In AE module the entailment relation is implemented in order to obtain the correct answer as it is proved it enhanced the accuracy of answer. The accuracy is defined as the average of the questions where the answer is found in the first rank.

Using an entailment based approach system seems to be a successful approach to extract answers for why questions. Its performance is good in the searching for candidate answers. EWAQ is described in two main phases: Pre-processing phase and processing phase

### A) Phase One: Pre-processing Phase

Question analysis:

- Removing stop words (irrelevant words).

    The stop words list of Abu-Elkhair [18] is employed in our system since removing the words of this words improves the retrieval precision and performance in Arabic language texts.

- Word Stemming.

    In EWAQ, the stemming approach suggested in [19] is used since it is more effective than other previous methods and obtained better average accuracy.

- Extracting the related words for each word in the Question and retrieved passages.

Using AWN, EWAQ extracts all the possible senses (words have semantic relations with this word) to each word in the question and passages since it increases the accuracy of IR systems.

### B) Phase Two: processing Phase

Implementing the entailment relation using AWN (with modification to be oriented to Arabic language). EWAQ measures the degree of entailment similarity between the why-question and passages retrieved by search engines like Ask, google and yahoo. EWAQ re-ranks retrieved passages according to their degree of entailment similarity.

Cosine directional similarity for textual entailment with the modification applied in our previous work [20] is used in this research as it has the highest computational performance for Arabic language. The authors of [20] made some modification to the entailment work applied in [21] to be oriented to Arabic language and get better results. The main modification in [20] are:

- More common words obtained by implementing the roots and semantics relations between the senses of the words.

- Based on experiments new the threshold values got.

The main implemented steps to measure the degree of entailment similarity in EWAQ are:

❖ Calculating the common words (c) between the question and each retrieved passage. (The common words between the question and the retrieved passage are words with the same root, and words that are related by semantic relations).

❖ Determining the length of each passage (m).

❖ Determining the length of why question (n).

❖ Certifying that m ≥n≥c.

❖ Appling the three methods equations are used by [21]:

$$\cos T(T;H) = \sqrt{c/m} \quad (1)$$

$$\cos H(T;H) = \sqrt{c/n} \quad (2)$$

$$\cos H \cup T(T;H) = \sqrt{4c^2/(n+c)(m+c)} \quad (3)$$

❖ Satisfying this primary condition $\cos H(T,H) \geq \cos H \cup T(T,H) \geq \cos T(T,H)$

❖ Checking the compulsory conditions to satisfy the entailment relation. The compulsory conditions are:

$\cos H \cup T - \cos T \leq \tau 1$ …(11)

$\cos H - \cos H \cup T \leq \tau$ …(12)

Max {cosT;cosH;cosH∪T} ≥ τ3 …(13)

❖ The thresholds used in this research are: τ1=0.095, τ2=0.2, τ3=0.5.

❖ When all entailment conditions are checked successfully, the degree of entailment similarity which we depend on it is cosH∪T.

Tables 1 and 2 show an example to the process of re-ranking retrieved passages, according to the entailment similarity, to the Arabic why question ( لماذا سميت زكاة الفطر بهذا الاسم" ,"lematha sommiat zakat elfeter behatha elesm?".

TABLE I. PASSAGES RETURNED BY GOOGLE FOR THE QUESTION
"لماذا سميت زكاة الفطر بهذا الاسم"

| Passage-No | Checking the conditions of entailment | The degree of entailment similarity | The passage |
|---|---|---|---|
| 1 | Succeed | 0.7872 | [Arabic passage] |
| 2 | Failed | ----- | [Arabic passage] |
| 3 | Succeed | 0.6493 | [Arabic passage] |
| 4 | Succeed | 0.8105 | [Arabic passage] |
| 5 | Failed | ----- | [Arabic passage] |
| 6 | Succeed | 0.7039 | [Arabic passage] |
| 7 | Succeed | 0.8345 | [Arabic passage] |

TABLE II. PASSAGES RE-RANKING BY EWAQ ACCORDING TO ENTAILMENT SIMILIARITY FOR THE QUESTION " لماذا سميت زكاة الفطر بهذا الاسم"

| Passages after re-ranking | Checking the conditions of entailment | The degree of entailment similarity | The passage |
|---|---|---|---|
| 7 | Succeed | 0.8345 | [Arabic passage] |
| 4 | Succeed | 0.8105 | [Arabic passage] |
| 1 | Succeed | 0.7872 | [Arabic passage] |
| 6 | Succeed | 0.7039 | [Arabic passage] |
| 3 | Succeed | 0.6493 | [Arabic passage] |
| 5 | Failed | ----- | [Arabic passage] |
| 2 | Failed | ----- | [Arabic passage] |

AE Module:

In the AE module, the task is to find the correct answer from the candidate answers within the relevant passages

In addition to implement Entailment relation to re-rank the relevant retrieved passages which improves the re-ranking retrieved passages, Entailment relation is implemented in AE module which adds another enhancement to why QA systems.

In order to obtain the correct-answers, the type of question included only why-questions, and the expected type is reason. We describe an AE module oriented to Arabic language for only why questions. In order to decide the final answer (mostly correct) we follow the following steps:

1. The top five passage re-ranked by EWAQ are selected as they gets the top values of entailment similarity.
2. If the retrieved passage consists of more than sentence, the text is split into sentences based on the existence of dot (.).
3. The same entailment algorithm implemented above is applied to extract the degree of entailment similarity between each sentence in each relevant passage and the why question.
4. Highest degree of entailment similarity is decided to be the correct answer.

V. EXPERIMENTS AND RESULTS

There are no available why Arabic question systems so that making a comparative study to determine the best one is hard, we have used search engines(google, yahoo and ask.com) as web resources to test our system. We conducted our evaluation on a set of 250 why questions selected by thirty professional persons and they are Arabic native speakers, with different fields (computer, religion, science, politic, history). Each field has fifty question. We store the selected questions with their correct answers which answered manually by human (thirty professional persons).

In order to obtain the accuracy of search engines (google, yahoo and ask) and EWAQ system. Each search engine is queried by the 250 questions and got the first seven retrieved passages. Using entailment metrics, we measure the degree of entailment similarity between each why question and its first seven retrieved passages by search engine. According to the degree of entailment similarity, the first seven retrieved passages are re-ranked. After re-ranking the retrieved passage, we apply the entailment metrics for each sentence in the re-ranked retrieved passage (if the passage consists of more than one sentence). The sentence with the maximum degree of entailment similarity, is assigned to be the correct answer.

Depending on the test set of 250 why questions with theirs correct answers (selected manually), we calculate the average of the questions where the answer is found in the first three rank to obtain the accuracy of each system. The results of accuracy were as follows in table 3:

TABLE III COMPARASION OF THE ACCURACY

| The system | The accuracy |
|---|---|
| ASK | 63.27 |
| GOOGLE | 66.19 |
| YAHOO | 61.48 |
| EWAQ | 68.53 |

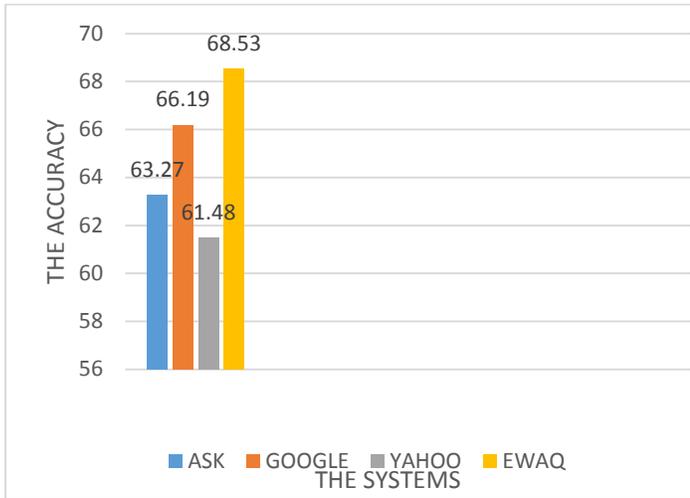

Figure 1. The obtained accuracy results of yahoo, google, ask and EWAQ systems

## VI. CONCLUSIONS

In this paper, it has been presented an approach for enhancing accuracy of Arabic why questions answering systems called EWAQ. The main objectives of EWAQ system is to improving the re-ranking passages relevant and retrieved by search engines (Yahoo, Google, Ask). The process of re-ranking the retrieved passages is based on measuring the degree of entailment similarity between the why questions and the retrieved passages.

To evaluate EWAQ, test set of 250 why questions with their correct answers answered manually is used. Since there are no Arabic QA systems implemented the why questions systems, search engines are used to compare the accuracy EWAQ with them.

Using entailment based similarity with some modification suited to Arabic language improves the accuracy of the why QA system. The overall accuracy with implementing the entailment similarity in EWAQ reached to 68.53%, where the values of accuracy reached to 66.19% for google, 63.27% for ASK, and the minimum accuracy is 61.48% for YAHOO. The experiments are very encouraging

## REFERENCES

[1] Diekema A., Yilmazel O., Chen J., Harwell S., Liddy E. and He L.," What do You Mean? Finding Answers to Complex Questions", In Maybury, M.T. (Ed.) New Directions in Question Answering. The MIT Press, pp. 141-152, 2004.

[2] Hirschman L. and Gaizauskas R.," Natural language QA: the view from here", Natural Language Engineering, vol. 7(4), pp.275-300, 2001.

[3] Voorhees E., "the TREC-8 QA Track Report, "In Proceedings of the Eighth Text Retrieval Conference (TREC-8), 2000.

[4] Ferret O., Grau B. and Huraults-Plantet M., Illouz G. Monceaux, L., Robba I., Vilnat A. "Finding an answer based on the recognition of the issue focus", In Proceedings ofTREC-10, 2001.

[5] Laurent D., Seguela P. and NègreCross S.,"Lingual QA using QRISTAL for CLEF 2006, Lecture Notes in Computer Science, Vol. 4730, pp. 339-350,2007.

[6] ASK website: http//:www.Ask.com- Last visited-April, 2015.

[7] Mohammed F., Nasser K. and Harb H., "A knowledge-based Arabic QA System (AQAS)," In Proceedings of ACM SIGART Bulletin, pp. 21-33, 1993.

[8] Hammo B., Abu-Salem H. and Lytinen S., "QARAB: A QA System to Support the Arabic Language". In Proceedings of the workshop on computational approaches to Semitic languages, pp. 55-65, Philadelphia, 2002.

[9] Benajiba Y., Rosso P. and Lyhyaoui A. "Implementation of the ArabiQA QA System's components", In Proceedings of Workshop on Arabic Natural Language Processing, 2nd Information Communication Technologies Int. Symposium, ICTIS, 2007.

[10] nooj website: http://www.nooj4nlp- Last visited-April, 2015.

[11] Brini W., Ellouze M. Mesfar S. and Belguith L. "An Arabic Question-Answering system for factoid questions",In Proceedings of IEEE International Conference on Natural Language Processing and Knowledge Engineering, 2009.

[12] Badawy O., Shaheen M. and Hamadene A. "ARQA High-Performance Arabic QA System", In Proceedings of Arabic Language Technology International Conference, pp. 129- 136, 2011.

[13] Akour M., Abufardeh S., Magel K. and Al-Radaideh Q.," QArabPro: A Rule Based Question Answering System for Reading Comprehension Tests in Arabic", American Journal of Applied Sciences, vol. 8 (6), pp. 652-661, 2011.

[14] Bdour W, Gharaibeh N., "Development of Yes/No ArabicQA System ", International Journal of Artificial Intelligence & Applications (IJAIA), Vol. 4(1), pp. 51-63, 2013.

[15] Dagan, I., Oren G., and Magnini B.," The Pascal recognizing textual entailment challenge", In Proceedings of the PASCAL Challenges Workshop on Recognizing Textual Entailment, 2005.

[16] Ramakrishnan G., Jadhav A., Joshi A., Chakrabarti, S. and Bhattacharyya P.," Question Answering via Bayesian Inference on Lexical Relations" , In Proceeding of  ACL Workshop Multilingual Summarization and Question Answering, pp. 1-10, 2003.

[17] Global WordNet website: http://globalwordnet.org/arabic-wordnet- Last visited-April, 2015.

[18] Abu-Elkhair I., "Effects of stop words elimination for arabic information retrieval: a comparative study", International Journal of Computing and Information Science, vol. 4(3), pp 119-133, 2006.

[19] Dilekh T., and Behloul A., "Implementation of a new hybrid method for stemming of Arabic text", International Journal of Computer Applications, vol.46 (8), 2012.

[20] AL-Khawaldeh F., Samawi V., "Lexical Cohesion and Entailment based Segmentation for Arabic Text Summarization (LCEAS)," The World of Computer Science and Information Technology Journal (WSCIT), Vol. 5(3), pp. 51, 60, 2015.

[21] Tatar D., Mihis A., and Lupsa D., "Entailment–based Linear Segmentation in Summarization", International Journal of Software Engineering and Knowledge Engineering vol. 19(80), pp. 1023–1038, 2009.